# SAF- BAGE: Salient Approach for Facial Soft-Biometric Classification - Age, Gender, and Facial Expression

Ayesha Gurnani[1,2]*   Kenil Shah[2]*   Vandit Gajjar[2,3]*   Viraj Mavani[1,2]*
Yash Khandhediya[2,4]*

[1]Erik Jonsson School of Engineering and Computer Science, The University of Texas at Dallas, USA
[2]Computer Vision Group, L. D. College of Engineering, India
[3]School of Engineering and Applied Science (SEAS), Ahmedabad University, India
[4]Dosepack LLC, Meditab Software Inc., India

```
{ang170003, viraj.mavani}@utdallas.edu, shah.kenil.484@ldce.ac.in
           vandit.gajjar@ahduni.edu.in, yashk@dosepack.com
```
*Authors are listed alphabetically and have contributed equally.

## Abstract

*How can we improve the facial soft-biometric classification with help of the human visual system? This paper explores the use of saliency which is equivalent to the human visual system to classify Age, Gender and Facial Expression soft-biometric for facial images. Using the Deep Multi-level Network (ML-Net) [1] and off-the-shelf face detector [2], we propose our approach - SAF-BAGE, which first detects the face in the test image, increases the Bounding Box (B-Box) margin by 30%, finds the saliency map using ML-Net, with 30% reweighted ratio of saliency map, it multiplies with the input cropped face and extracts the Convolutional Neural Networks (CNN) predictions on the multiplied reweighted salient face. Our CNN uses the model AlexNet [3], which is pre-trained on ImageNet.*

*The proposed approach surpasses the performance of other approaches, increasing the state-of-the-art by approximately 0.8% on the widely-used Adience [28] dataset for Age and Gender classification and by nearly 3% on the recent AffectNet [36] dataset for Facial Expression classification. We hope our simple, reproducible and effective approach will help ease future research in facial soft-biometric classification using saliency.*

## 1. Introduction

Since Deng *et al.* [4] and Russakovsky *et al.* [5] has proposed large-scale image database and classification challenge ImageNet, with the help of Deep Convolutional Neural Networks (D-CNN), much progress has been made in the field of computer vision. Several novel architectures have been proposed and reduced the error-rate for the image classification tasks [3, 6, 7, 8]. The use of deep learning not only has revolutionized the field of computer vision but in many other research directions, it has made progress, e.g. in the well-known Chinese game named as Go [9], in Audio classification [10], and Natural Language Processing [11].

Human face analysis and soft-biometric classification, on the other hand, has gained more popularity after AlexNet has been introduced by Krizhevsky *et al.* [3]. Such facial soft-biometric include Age, Gender and Facial Expression are a topic of interest among many computer vision researchers. Introduction of deep learning to this domain has replaced the need for handcrafted facial attributes and data pre-processing schemes. D-CNN models have been not only successfully applied to Computer Vision, but also for the Cognitive Vision tasks, especially s*aliency* detection [12, 13, 14]. *Saliency* is fundamentally an intensity map where higher intensity signifies regions, where a general human being would look, and lower intensities mean decreasing level of visual attention. It's a measure of visual attention of humans based on the content of the image. It is still an open problem when considering the MIT Saliency Benchmark [15].

In the previous five years, considering Age, Gender and Facial Expression soft-biometric classification accuracies increased rapidly on several benchmarks. However, in some unconstrained environments, i.e. in Adience benchmark some facial images have low-resolution and different head poses. In this type of facial images, soft-biometric classification is still facing challenges to achieve competitive results. Thus in order to solve this problem, inspired by the recent progress in the domain of Computer and Cognitive Vision, we explore the use of s*aliency* to classify Age, Gender and Facial Expression on facial images. Based on the researcher's previous work and above motivation, our work for the facial soft-biometric classification task is as follows:

Our approach SAF-BAGE uses three modules. Face module uses an off-the-shelf face detector proposed by Mathias *et al.* [2] to obtain the location of the face in the test image. Then, the margin of the detected B-Box increases by 30% to get the complete face. *Saliency* module is followed by Face module which uses the ML-Net. ML-Net is recognized as one of the dominant models for *saliency* prediction, so *saliency* map is obtained from the network. After obtaining the map, we multiply the input face and salient face map (with 30% reweighted ratio) to get the multiplied reweighted salient face. The final module is a Convolution module, which is for soft-biometric classification. We use the pre-trained AlexNet architecture. We fine-tune the AlexNet on two known benchmarks: Adience and AffectNet: A Database for Facial Expression in the wild. Fine-tuning the CNN on large-scale training images is an essential step to gain benefit from the representation power of the CNN.

We perform an experiment on Adience and AffectNet, to show the effectiveness of our approach.

Our main contributions of this paper are as follows:

- ➢ We present our approach - SAF-BAGE, which surpasses the current state-of-the-art on Adience and AffectNet dataset for Age, Gender, and Facial Expression classification has the ability to learn the salient features after computing *saliency* in order to precisely classify facial soft-biometrics.

- ➢ We demonstrate that the use of 30% *saliency* reweighted ratio would help the model to converge faster because the multiplied reweighted salient image would have less variance as compared to normal images in training.

- ➢ We present a semi-supervised approach, in which we did not use facial images explicitly to train the *saliency* detection algorithm. We used the pre-trained *saliency* model, trained on SALICON dataset, which is considered as a large-scale *saliency* benchmark. This dataset contains different image classes including human faces.

- ➢ We show that using Class Activation Maps, for the facial images, Eyes, Nose, and Face Regions are the dominant facial attributes, which helps the model for soft-biometric classification.

The rest of this paper is as follows. In Section 2 we discuss the related work about Salient Object Detection and Age, Gender and Facial Expression classification. In Section 3, we describe our approach - SAF-BAGE, its modules, and the implementation details. Experiments and results are briefly described and shown in Section 4. Section 5 discusses the effectiveness of the work, focuses on the possible future work, and concludes the paper.

## 2. Related Work

### 2.1. Salient Object Detection

Over the last decade, much progress has been reported in *saliency* detection. Several supervised and unsupervised *saliency* detection algorithm/methods have been proposed under different theoretical models [16, 17, 18]. A *saliency* model proposed by Itti *et al.* [16] linearly combines image attributes including color, intensity to recognize local prominence. However, this method tends to enlighten the salient pixels and loses object information. To characterize the spatial arrangement of image regions, Zhu *et al.* [19] proposed a background measurement technique. Cheng *et al.* [20] have addressed *saliency* detection based on the global region contrast, which simultaneously considers the spatial coherence across the regions and the global contrast over the image. The methods and approaches that consider only local contexts tend to recognize high-frequency information and suppress the region inside salient objects. In an image, to compute the color difference between each pixel with respect to its mean, Achanta *et al.* [18] has proposed a method to estimate *saliency*. The work proposed by Liu *et al.* [21] uses both local and global set of attributes, which are integrated by a random field to generate a *saliency* map. Although compelling improvements have been made, most of this methods and approaches integrate hand-crafted features to create *saliency* map. Thus, in order to predict *saliency* map and learn features automatically, we have used the ML-Net proposed by Cornia *et al.* [1]. The proposed model is used to calculate low and high-level features from the input image. For generating *saliency*-specific feature maps, the extracted features are fed to an encoding network, which learns a feature weighting function. Section 3 discusses the model in detail.

### 2.2. Age, gender and facial expression classification

A recent and detailed survey of the facial soft-biometric: Age, Gender and Facial Expression classification can be found in the work of [22] and [23]. Here, we quickly discuss the appropriate methods/approaches. Age, Gender and Facial Expressions are one of the key facial soft-biometric, which plays a significant role in our social interactions. In the early 90's many approaches [24, 25, 26] have been proposed for this soft-biometrics classification. These approaches basically use the handcrafted features for classification tasks, i.e. support vector machines. The major revolution has begun since, the D-CNN have been introduced for image classification task, (i.e AlexNet proposed by Krizhevsky *et al.* [3]). Then onwards many researchers have used the deep learning approaches for the classification task. Levi *et al.* [27] adopted the AlexNet architecture and have used for the very first time three-layer CNN for Adience benchmark [28] for Age and Gender classification. Alizadeh *et al.* [29] have developed CNN for a Facial Expression classification on a gray-scale dataset provided by Kaggle website.

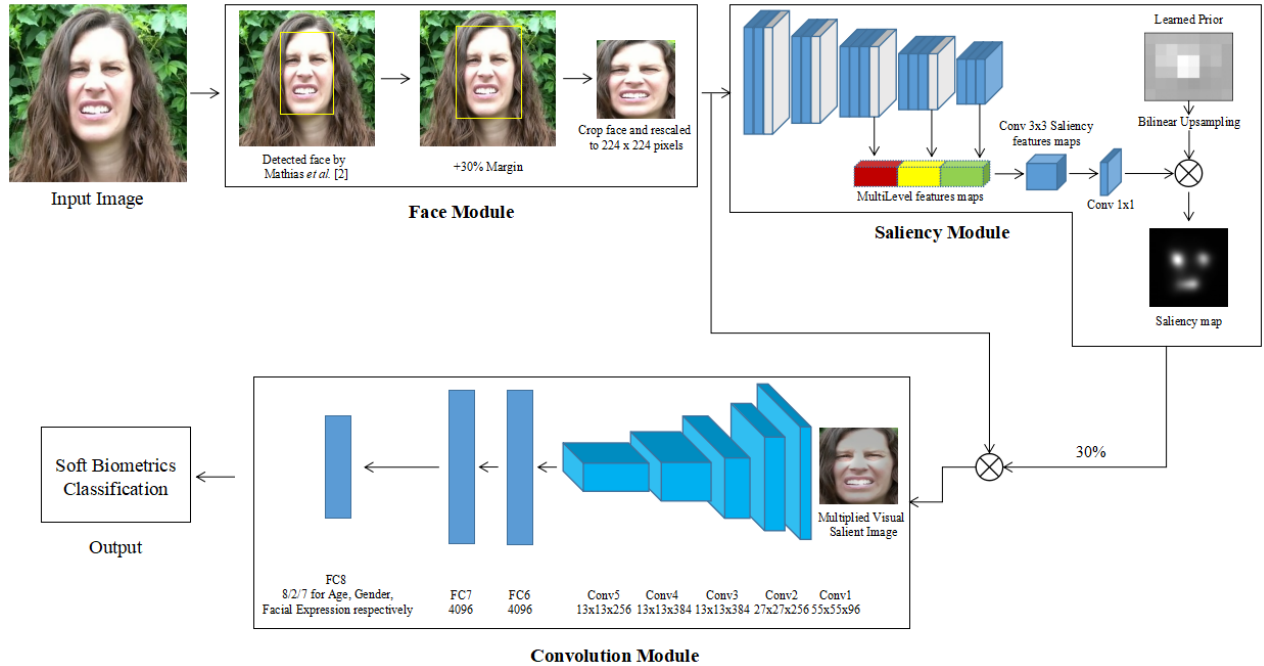

Figure 1: **Overview of the proposed approach - SAF-BAGE for facial soft-biometric classification.** We first apply the face detector to the image for face detection in Face module. Detected face's B-Box margin is increased to 30% and cropped. We then obtain the *saliency* map from the cropped face using ML-Net [1] in *saliency* module. The cropped face and 30% reweighted salient map is multiplied to obtain the multiplied reweighted salient face. This face is then fed into a classification network (fine-tuned AlexNet [3]) for the final soft-biometric classification in Convolution module. (Best viewed in color and magnification.)

Afterward, numerous approaches/methods for facial soft-biometric classification have been proposed by researcher's i.e. Rothe *et al.* [30] proposed DEX: Deep EXpectation of Apparent Age From a Single Image method for Age classification using the ensemble of 20 networks on the cropped faces of IMDB-Wiki benchmark. Jung *et al.* [31] have proposed a novel joint fine-tuning method for Facial Expression classification. Liao *et al.* [32] have presented an approach using local deep neural networks. Authors have proposed to use the nine overlapping patches per image instead of an entire facial image, which significantly reduces the training time. Ranjan *et al.* [34] have presented a multi-task CNN-based method for classification of Age, Gender and Smile for facial image. The method detects the face in the test image, extract facial key-points and determine its Age and Gender from a less constrained environment. The method also assigns an identity descriptor to each face which can be further used for face detection and verification. The proposed method was trained on several major facial benchmarks including CASIA, MORPH, IMDB-WIKI, Adience, CelebA, and AFLW with almost 1 million training samples. Dehghan *et al.* [33] has proposed a fully automatic Age, Gender and Facial Expression classification system with a backbone of several D-CNN and achieves the state-of-the-art method on several benchmarks. Unfortunately, there is no specific details are given about the backbone architecture in the system that has been used. We show our proposed approach outperforms the results they report on the Adience benchmark.

## 3. Our proposed approach: SAF-BAGE

The detailed illustration of our approach - SAF-BAGE is shown in Figure 1. Firstly, we propose a simple yet effective approach for facial soft-biometric classification using *saliency*. The input image is given to face module in which off-the-shelf face detector [2] detects the precise face and gives the B-Box coordinates. We use the obtained B-Box coordinates and increases the margin of B-Box by 30% to get the complete face and then crop it. The *saliency* module is followed by a Face module, in which the cropped face is passed from ML-Net [1] to find the *saliency* map. The product of input crop face and reweighted *saliency* map (30%) is fed to AlexNet, which is pre-trained on ImageNet and finetuned on the datasets of Adience and AffectNet. Next subsections provide process details of each module.

**Face Module:** For accurate facial soft-biometric classification, one of the essential and first steps is accurate and robust face detection. There are several face detection modules available. One of the major reasons to use the face detector proposed by Mathias *et al.* [2] in our approach is that it has achieved the Average Precision (AP) of 89.63 on Pascal Face and AP of 97.14 on Annotated Faces in the Wild (AFW), which are recognized as extreme wild datasets for face detection. To obtain the precise location of the face and improve the classification accuracy, we extend the margin of detected B-Box by 30%. If the face already covers most of the image, we pad with the last pixels at the border. Supporting the claim of [30], adding this context helps the classification accuracy. The resulting image is then re-scaled to 224 x 224 pixels and used as an input to the *saliency* module.

**Saliency Module:** Face module produces the 224 x 224 pixels cropped face. The *saliency* module uses the ML-Net [1] for *saliency* prediction. The ML-Net architecture is built on the VGG-16 [6] model. It is composed of three parts: Input face image, a CNN which extracts low, medium and high-level features; and last an encoding network, which learns a feature weighting function and produces *saliency* map. In the encoding network, the feature maps at three different locations are taken: Third Pooling Layer, that of the last pooling layer, and the output of the last Convolutional layer. These feature maps share the same spatial size and are concatenated. A final 1 x 1 convolution learns to weight the importance of each *saliency* feature map to produce the final predicted feature map.

In our case, we have used SALICON pre-trained weights for *saliency* prediction. The primary reason to choose ML-Net by Cornia *et al.* [1] is that it has outperformed all other models on the SALICON Dataset [35] and also performs better on the MIT Saliency Benchmark [15]. After obtaining *saliency* map, the product of input cropped face resulted from Face module and salient face map with 30% reweighted ratio is used as an input to the Convolution module for final classification.

**Convolution Module:** The product is multiplied reweighted salient face, which can be seen in Figure 1 (input of convolution module). We fed this face to our fine-tuned AlexNet. One of the motives to choose AlexNet as a classification network is that it has achieved the error rate of 15.3% on ImageNet test set, which was the winning entry in ImageNet competition back in 2012. Also, the approach contains two other modules, so to make an end-to-end framework possible with computing constraint, we used AlexNet as a core module for prediction. The AlexNet contains eight learned layers - five convolutional and three fully-connected. In our case, the output of the last fully-connected layer is fed to an 8/2/7-way softmax which produces a distribution over the given class labels. The ReLU (Rectified Linear Units) is applied to the output of every convolutional and fully-connected layer. For the classification task, the AlexNet is fine-tuned on Adience and AffectNet datasets for Age, Gender and Facial Expression. The final output produces the corresponding Age, Gender, and Facial Expression label according to the input image.

### 3.1. Dataset overview

The Adience [28] and recently released AffectNet [36] are currently recognized as wild and large-scale datasets for facial soft-biometric classification. For both the datasets, the details are below.

**Adience:** The entire Adience dataset contains approximately 26,000 images of 2284 subjects. The benchmark consists of images which are taken from smart-phone devices and uploaded to Flickr. Due to this capturing, uploaded images were without filtering, reflecting real-world facial images. Therefore, Adience images are having extreme variations in head pose, various lighting quality, and noise.

**AffectNet:** AffectNet dataset is recently released dataset and considered as the largest database for Facial Expression. It contains more than 10,00,000 facial images of different faces from the Internet by querying three major search engines (Google, Bing, and Yahoo) using 1250 emotion-related keywords in six different languages. The images contain at least one face with its extracted facial landmark points. The average image resolution of faces in AffectNet is 425 x 425 with STD of 349 x 349 pixels. However, the author has not made the test set publicly available, and urge researchers to use the validation set as a baseline. Therefore, Train, Val, and Test split are carried on a total of 2,87,400 images having 7 expressions - Neutral, Happy, Sad, Surprise, Fear, Disgust, and Anger. Table 1 shows the Train, Val, and Test split across images.

| Expression | Train | Val | Test |
| --- | --- | --- | --- |
| Neutral | 59900 | 14973 | 500 |
| Happy | 107532 | 26883 | 500 |
| Sad | 20367 | 5092 | 500 |
| Surprise | 11272 | 2818 | 500 |
| Fear | 5102 | 1276 | 500 |
| Disgust | 3042 | 761 | 500 |
| Anger | 19906 | 4976 | 500 |
| Total | 227121 | 56779 | 3500 |

Table 1: **Train, Val, and Test split** of the AffectNet dataset for Facial Expression classification.

### 3.2. Performance metric

The performance measurement for both Age and Gender classification is tested using a standard five-fold, subject-exclusive cross-validation protocol, defined by Eidinger *et al.* in [28].

We follow the evaluation for Facial Expression classification reported by Mollahosseini *et al.* in [36]. Human annotator agreement for Facial Expression classification on AffectNet is slightly more over 60%. We consider the weighted-loss approach for evaluation. In classification, the images of Disgust category achieve an acceptable accuracy but didn't perform well in Neutral class. This is because the network was not penalized enough for miss-classifying examples from these classes. For the Facial Expression classification accuracy, it is defined in a multi-class manner in which the number of correct predictions is divided by the total number of images in the test set.

### 3.3. Implementation details

The proposed approach uses the pre-trained weights of SALICON [35] and ImageNet [3] for *saliency* prediction and soft-biometric classification respectively. The fine-tuning is accomplished on a workstation with Intel Xeon core processor and accelerated by NVIDIA TitanX 12 GB GPU. All experiments run in Tensorflow 1.6 [41].

**Age and gender fine-tuning:** It is a practice in deep learning to augment the training images for improvement in performance and accurate classification. E.g. fine-tune AlexNet with 19487 images performed using a five-fold cross-validation for Age and Gender classification may result in over-fitting, which we avoid using data augmentation. Thus, each image is horizontally flipped, rotated with 2 angles {3º, -3º}. This generated 4 times more data. The last three layers of AlexNet namely { fc6, fc7, and fc8} are fine-tuned for Age and Gender classification. It is trained for 30 epochs with the learning rate of 0.001, dropout set to 0.4 and effective batch size of 128.

**Facial expression fine-tuning:** The training image size is very high for Facial Expression dataset, thus use of data augmentation is not needed for our fine-tuning process. The fine-tuning is accomplished by the last 4 layers {Conv5, fc6, fc7, and fc8} of AlexNet. The model is fine-tuned for 40 epochs. The learning rate is set to 0.001 and divided by 10 after 30 epochs. The effective training batch size is set to 128 and dropout set to be 0.35.

### 4. Experiments and results:

In this section, we outline the experiments that demonstrate the benefit of *saliency* in soft-biometric classification.

Our experiment compares with and without *saliency* module for soft-biometric classification. In without *saliency* module, the hyper-parameters are all same as with *saliency* module in training time.

### 4.1. Age and gender classification

Table 2 and Table 3 presents our results for Age and Gender classification with and without *saliency* module, while Table 4 provides a confusion matrix for Age classification with *saliency* module. For Age classification, we measure the accuracy when the approach predicts the exact Age-group.

| Method | Accuracy |
| --- | --- |
| Eidinger *et al.* [28] | 45.1 ± 2.6 % |
| Levi *et al.* [27] (Single Crop) | 49.5 ± 4.4 % |
| Levi *et al.* [27] (Multi Crop) | 50.7 ± 5.1 % |
| Ours (Without *saliency* Module) | **52.2 ± 3.9 %** |
| Qawaqneh *et al.* [37] | 59.9 % |
| Dehghan *et al.* [33] | 61.3 ± 3.7 % |
| Ours (With *saliency* Module) | **62.11 ± 3.2%** |

Table 2: **Age classification accuracy results** on Adience benchmark with our approach, including without the *saliency* Module. Listed are the mean accuracy ± standard error over all Age categories. Our results are marked in bold.

| Method | Accuracy |
| --- | --- |
| Eidinger *et al.* [28] | 77.8 ± 1.3 % |
| Liao *et al.* [32] | 78.63 % |
| Hassner *et al.* [38] | 79.3 ± 0.0 % |
| Ours (without *saliency* module) | **83.4 ± 1.6 %** |
| Levi *et al.* [27] | 85.9 ± 1.4 % |
| Levi *et al.* [27] | 86.8 ± 1.4 % |
| Dehghan *et al.* [33] | 91 % |
| Ours (with *saliency* module) | **91.8 ± 1.2 %** |

Table 3: **Gender classification accuracy results** on Adience benchmark with Our approach, including without the *saliency* Module. Our results are marked in bold.

From Table 4, we noticed that the 0-2-year-old Age contain different facial features that enable the classifier to differentiate this Age-group very easily. While 38-43-year-old Age-group is quite difficult for the classifier to predict and has the lowest accuracy of 27.3%. This Age-group is adjacent to labels of 25-32-year-old Age-group. A performance comparison between the proposed approach (With and Without *saliency* module) and the state-of-the-art methods are shown in Table 2. It also provides comparison with [28, 29, 37, 33].

On the other side, same for Gender classification, we measure the accuracy when the true positive comes in the prediction. Table 3 compare performance with [28, 27, 32, 38, 33]. With the help of *saliency*, the approach achieves the state-of-the-art on the task of Age and Gender classification with accuracy of 62.11 ± 3.2% and 91.8 ± 1.7% respectively.

| Age Group | 0-2 | 4-6 | 8-13 | 15-20 | 25-32 | 38-43 | 48-53 | 60- |
|---|---|---|---|---|---|---|---|---|
| 0-2 | **92.1** | 4.2 | 0.0 | 1.3 | 0.0 | 2.4 | 0.0 | 0.0 |
| 4-6 | 22.2 | **70.2** | 5.8 | 3.1 | 0.0 | 0.0 | 0.4 | 0.1 |
| 8-13 | 3.6 | 12.6 | **52.8** | 13.2 | 11.7 | 3.8 | 2.3 | 0.0 |
| 15-20 | 1.4 | 0.3 | 13.7 | **36.3** | 42.6 | 3.7 | 1.2 | 0.8 |
| 25-32 | 0.2 | 0.0 | 0.1 | 4.7 | **88.4** | 3.8 | 2.3 | 0.5 |
| 38-43 | 0.0 | 0.4 | 0.7 | 3.8 | 49.8 | **29.2** | 13.8 | 2.3 |
| 48-53 | 0.8 | 0.0 | 0.0 | 0.9 | 2.6 | 18.3 | **47.8** | 29.6 |
| 60- | 0.0 | 0.3 | 0.4 | 2.8 | 1.7 | 3.5 | 11.2 | **80.1** |

Table 4: **Age classification confusion matrix** on the Adience benchmark with our approach (with *saliency* Module).

|  | Neutral | Happy | Sad | Surprise | Fear | Disgust | Anger |
|---|---|---|---|---|---|---|---|
| Neutral | **62.4 (312)** | 6.6 (33) | 7.8 (39) | 7.4 (37) | 3.2 (16) | 3.8 (19) | 8.8 (44) |
| Happy | 3.8 (19) | **76.4 (382)** | 2.0 (10) | 5.0 (25) | 1.2 (6) | 8.4 (42) | 3.2 (16) |
| Sad | 21.2 (106) | 2.8 (14) | **64.4 (322)** | 3.4 (17) | 4.0 (20) | 3.6 (18) | 0.6 (3) |
| Surprise | 7.8 (39) | 5.0 (25) | 3.0 (15) | **73.0 (365)** | 8.6 (43) | 1.4 (7) | 1.2 (6) |
| Fear | 2.2 (11) | 5.2 (26) | 6.0 (30) | 11.6 (58) | **68.4 (342)** | 4.8 (24) | 1.8 (9) |
| Disgust | 5.0 (25) | 3.4 (17) | 1.4 (7) | 3.8 (19) | 18.2 (91) | **66.2 (331)** | 2.0 (10) |
| Anger | 6.0 (30) | 1.6 (8) | 5.8 (29) | 5.6 (28) | 5.8 (29) | 12.4 (62) | **62.8 (314)** |

Table 5: **Facial Expression classification confusion matrix** on the AffectNet dataset with our approach (with *saliency* module). The Number in the bracket shows the number of frames.

## 4.2. Facial expression classification

The confusion matrix for the Facial Expression classification with *saliency* module is provided in Table 5.

| Method | Accuracy |
|---|---|
| Söderberg *et al.* [40] | 48.58 % |
| Hewitt *et al.* [39] | 57.8 % |
| Ours (without *saliency* module) | **59.2 %** |
| Mollahosseini *et al.* [36] | 64.53 % |
| Ours (with *saliency* module) | **67.65 %** |

Table 6: **Facial Expression classification accuracy results** on AffectNet benchmark with Our approach, including without *saliency Module*. Our results are marked in bold.

The weighted-loss approach classified the samples of Surprise (73.0%) and Happy (76.4%) categories with an acceptable accuracy but did not perform well in Neutral (62.4%), Anger (62.4%), and Sad (64.4%). We observed that Happy Expression has the highest rate of correct classification (76.4%), while Neutral (62.4%) has the lowest accuracy and it often being confused with Disgust. In Table 6, we compare the performance with [36, 39, 40]. We achieve state-of-the-art by more than 3.12 % compare to reported accuracy in the work of Mollahosseini *et al.* [36].

Figure 2 shows the main results of our approach. The Input crop face, it's *saliency* map, multiplied reweighted salient face and final soft-biometric classification of the input image.

## 4.3. Ablation experiments

We run some ablations to analyze our SAF-BAGE approach. With heavy computation needed in Facial Expression classification, we performed our ablations on Gender classification task only.

**Choice of classification network:** To test the influence of classification network on the proposed approach, we fine-tune VGG-16 network [6] (Only last fully-connected layer due to computing power) and GilNet [27]. The VGG-16 network is about 1.6% better than the AlexNet, while the GilNet shows a comparable result with slightly less accuracy of 89.8 ± 1.3%.

**Reweighted saliency ratios study:** In our approach, one changeable parameter is the use of 30% reweighted *saliency* ratios. We try different *saliency* ratios to check the performance of our approach. Therefore, we fine-tuned the AlexNet network with 10%, 50%, 70%, and 90% multiplied reweighted salient face images. Figure 3 shows the training curve - Loss vs. Epoch, while Figure 4 shows the validation curve - Validation Accuracy vs. Epoch. From both the figures it can be clearly seen that the 90% reweighted *saliency* ratio helps the model to converge faster as the image has less variance, however, the performance of that model is very poor, as the model is not able to learn the salient features. Here, we noticed that higher reweighted

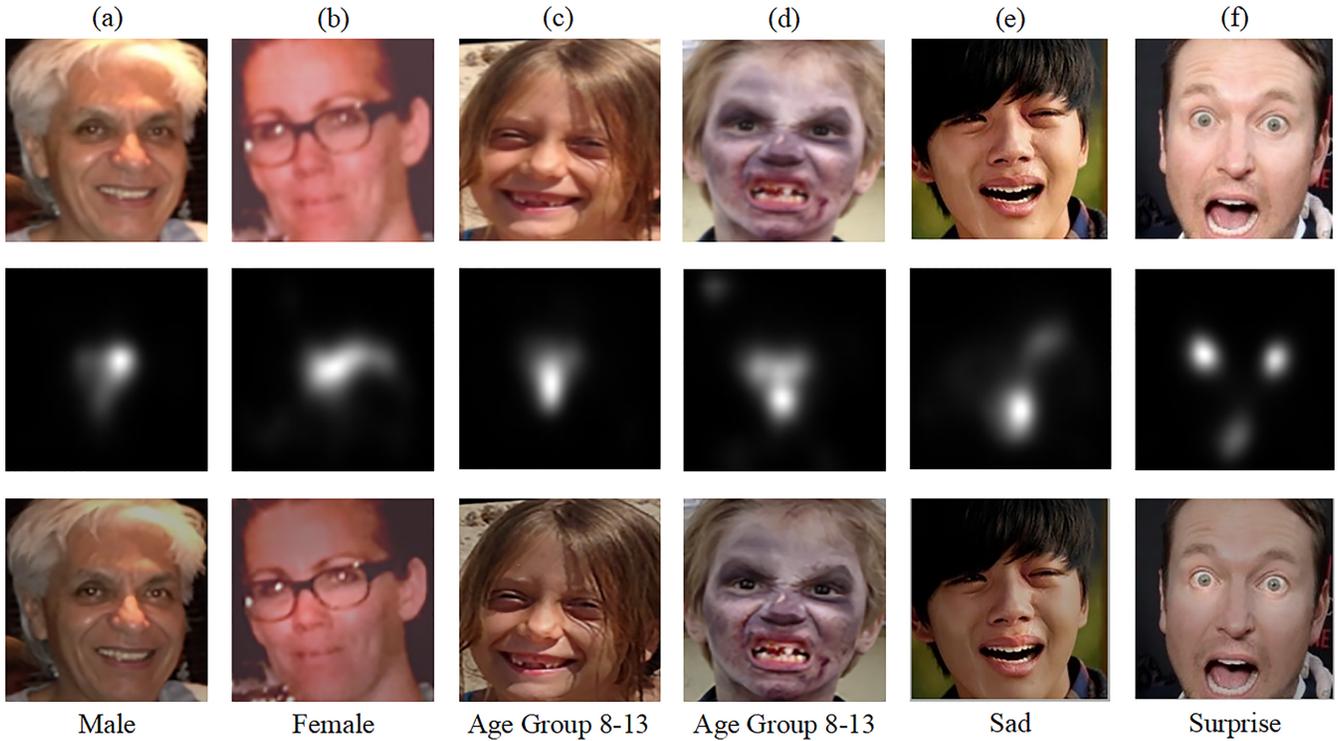

Figure 2: **Example results for images from the Adience and AffectNet test sets.** Per image, we show *saliency* map obtained from the ML-Net [1] model and multiplied with the input cropped image for the classification. Here, (a) and (b) are images from the Adience dataset for Gender classification, (c) and (d) are images from the Adience dataset for Age-group classification and (e) and (f) are images from the AffectNet dataset for Facial Expression classification. Also in the example results, we can see that Salient maps are mainly focused on **Eyes, Nose, and/or near to Mouth regions**. This shows that these three are the dominant salient attributes, which are helping the model to classify such facial soft-biometric like Age, Gender and Facial Expression.

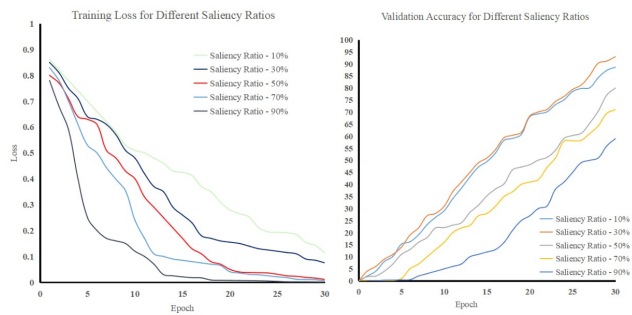

Figure 3: **Training Loss and Validation Accuracy visualization for different *saliency* ratios.** Left: We show training curves using different *saliency* ratios - 10%, 30%, 50%, 70%, 90%. This verifies that using the *saliency* it helps the model to converge faster. Right: We show validation accuracy using different *saliency* ratios on Adience dataset for gender classification. This verifies that the fastest converged model (90% reweighted *saliency* ratio) not helps to improve the performance. With both the graphs, we verify that the choice of 30% reweighted *saliency* ratio helps the model to converge fast, and also improve the performance. (Best viewed in color and magnification)

*saliency* ratio helps the model to converge faster but leads to poor performance. Thus, the choice of 30% reweighted *saliency* ratio confirms the effectiveness of our approach.

## 5. Discussion and conclusion

Our proposed approach SAF-BAGE shows a new direction for training a neural network, that using multiplied reweighted salient image, the performance could be further improved for facial soft-biometric classification. Also from the Figure 2, it can be seen that the salient regions - Eyes, Nose, and Mouth, these three are the dominant attributes which help the model to classify facial soft-biometric. Further, using Class Activation Maps (CAM) [42], we also verify these dominant regions are playing a significant role for classification. In CAM, for AlexNet [3] we remove the fully-connected layers before the final output and replace it with Global Average Pooling (GAP) followed by a softmax layer. We can identify the importance of the image regions by projecting back the weights of the output layer on the feature maps, by giving this simple structure. Figure 4, shows the heat-map from the last Convolutional layer for AffectNet database images.

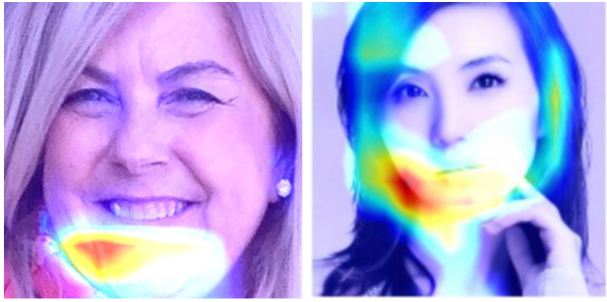

Figure 4: **Class Activation Map.** In two images, we can see that for the classification, Hair, Face, and Mouth regions are activated.

*Q: why multiplied reweighted salient image instead of a normal image?*

We just want to slightly eliminate the background, rather than steer the focus. Normal images sometimes create abrupt edges that confuse the filters during the training. When using salient images, low-mid frequency information is still well preserved. One can still identify the content of the image, i.e. Gender, Facial Expression from the salient image. In a multiplied reweighted salient image, outside the particular region, and leaving the high-frequency information at the particular region will help to provide global context of the image, as well as keeping the details intact at the region of interest, which will help in the task of facial soft-biometric classification. We report that by providing the multiplied reweighted salient image to a neural network, it is easy for the network to converge faster and extract the features.

Another promising future extension of this work would be to implement this approach for Face detection and Face retrieval in unconstrained surveillance. The architectural variant is possible in our approach, thus other architectures such as VGG-16, ResNet, DenseNet may help to improve the classification performance. Also, through the multiplication with *saliency* map, if the salient regions are clipped and passed through the network, it could significantly improve the performance.

Thus, three major conclusions can be made from our experimental results. First, using human vision system - *saliency*, the facial soft-biometric classification performance surpassed the current state-of-the-art for Age, Gender and Facial Expression classification. Second, with multiplied reweighted salient face image helps the model to converge faster and leads to improvement in the classification performance. Third, with our proposed approach SAF-BAGE, we have observed that salient regions - Eyes, Mouth, and Noise are playing an important role in the soft-biometric classification. Finally, the simplicity of our method implies that with wild and more number of training data may well be capable to improve the performance greatly beyond reported here. We hope our simple, reproducible, and effective approach will help ease future research in facial soft-biometric classification using *saliency*.

**Acknowledgments:** We are thankful to the anonymous reviewers for their valuable comments and suggestions due to which the paper was improved. We gratefully acknowledge the support of NVIDIA Corporation with their donation of the Titan XP GPU. The authors would also like to thank Bansi Gajera, Dhwani Mehta, Jeet Jivrajani, Kanan Vyas, Param Rajpura, Priya Mehta, Raj Bhensadadia, and Vishwa Saparia for insightful discussions; Maharshi Doshi and Vishwa Shah for their help with the manuscript.